# Intelligent humanoids in manufacturing to address worker shortage and skill gaps: Case of Tesla's Optimus

**Ali Ahmad Malik**
Oakland University
Michigan, United States
Email: aliahmadmalik@oakland.edu

**Tariq Masood**
Strathclyde University
Glasgow, Scotland, United Kingdom
Email: tariq.masood@strath.ac.uk

**Alexander Brem**
Stuttgart University
Stuttgart, Germany
Email: alexander.brem@eni.uni-stuttgart.de

**Pre-print**

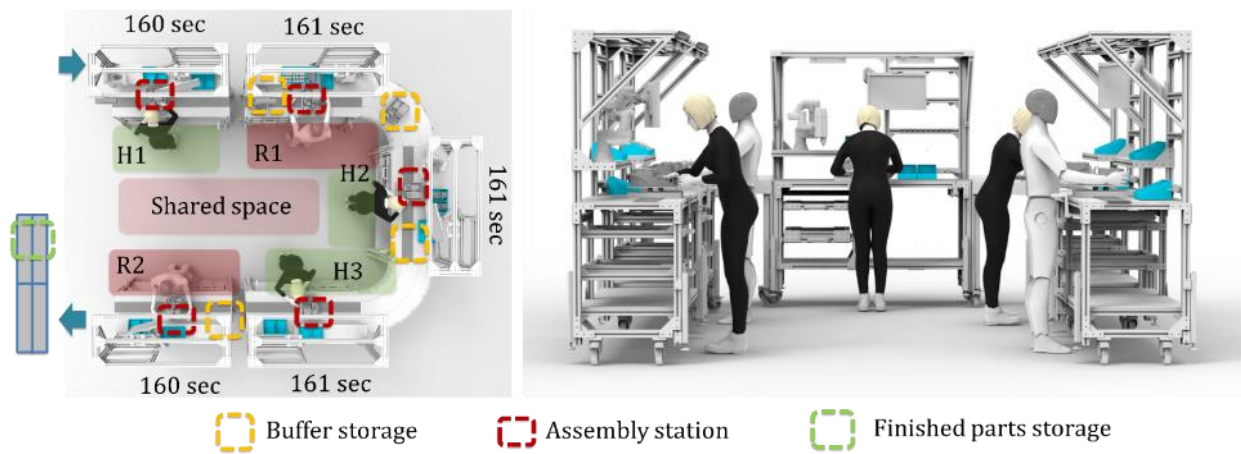

**Highlights**

1. Humanoid robots are gaining traction but their implication in the manufacturing practices especially for industrial safety standards and lean manufacturing practices have been minimally addressed.
2. This article proposes a framework to integrate humanoids for manufacturing automation and also presents the significance of safety standards of human-robot collaboration
3. Simulation based digital twin is proposed for verification and validation of humanoids-based manufacturing systems.

# Intelligent humanoids in manufacturing to address worker shortage and skill gaps: Case of Tesla's Optimus

**Ali Ahmad Malik**
Oakland University
Michigan, United States
Email: aliahmadmalik@oakland.edu

**Tariq Masood**
Strathclyde University
Glasgow, Scotland, United Kingdom
Email: tariq.masood@strath.ac.uk

**Alexander Brem**
Stuttgart University
Stuttgart, Germany
Email: alexander.brem@eni.uni-stuttgart.de

*Abstract:*

*Technological evolution in the field of robotics is emerging with major breakthroughs in recent years. This was especially fostered by revolutionary new software applications leading to humanoid robots. Humanoids are being envisioned for manufacturing applications to form human-robot teams. But their implication in the manufacturing practices especially for industrial safety standards and lean manufacturing practices have been minimally addressed. Humanoids will also be competing with conventional robotic arms and effective methods to assess their return on investment are needed. To study the next generation of industrial automation, we used the case context of Tesla's humanoid robot. The company has recently unveiled its project on an intelligent humanoid robot named 'Optimus' to achieve an increased level of manufacturing automation. This article proposes a framework to integrate humanoids for manufacturing automation and also presents the significance of safety standards of human-robot collaboration. A case of lean assembly cell for the manufacturing of an open source medical ventilator was used for human-humanoid automation. Simulation results indicate that humanoids can increase the level of manufacturing automation. Managerial and research implications are presented.*

### 1. Introduction

On the one hand, the world is facing a workforce shortage due to aging humanity in industrialized countries [1][2]. In addition, there is a great resignation and poor perception of manufacturing jobs among young people [3]. At the same time, manufacturing, being a major contribution of any economy for national GDP, still has a high reliance on human workforce [4]. The potential of manufacturing automation is limited and several manufacturing operations are still more prone to human labor. Typical examples are assembly lines where humans are needed to perform boring, repetitive, and sometimes dangerous tasks [5]. On the other hand, assembly is not easy to automate with conventional robotics [6].

Technological advancements [7], their anticipated impact [8] [9], and supportive government strategies [10] forecast a bright future for the manufacturing sector. But the way to achieve this future of manufacturing excellence is not straightforward for most enterprises [11]. Besides the challenges of effectively leveraging digitalized technologies, an important challenge is skill shortage [2]. It is predicted that by 2030, 2.4 million jobs will remain unfilled due to a lack of necessary skills in the U.S. alone [12].

Automation helps to reduce human effort [13]. Industrial robots constitute the backbone of present-day industrial automation [14]. These robots are available in varying configurations, such as articulated, SCARA, delta, mobile robots, etc., to perform a variety of tasks [15]. Besides 3.5 million robot installations worldwide, and with a global robot density of 126 units for every 10,000 employees [16], humans are still vital to perform many of the boring, repetitive, and sometimes dangerous tasks at manufacturing floors [6]. The fact [17] that humans are the most flexible resource on a manufacturing floor while robots are the opposite, defines the reason for gaining manufacturing flexibility through human resources. Humans possess high-level skills of mobility in uncertain environments [18], the ability to perform a variety of tasks [19] as well as to support other humans [20].

Humanoid robots mimic humans [21]. They represent human-like characteristics in their mechanical architecture [22], dexterity, and cognition skills [23]. Although humanoids remained attractive in fiction and movies [24] but received limited attention for manufacturing automation [22]. Some efforts to realize humanoid robots include ASIMO by Honda, NextAge by Kawada Robotics, Robonaut by NASA, and Atlas by Boston Dynamics [23]. However, efforts remained in the development phases and never made the real use for which they were aimed.

Tesla (a manufacturer of electric vehicles) unveiled its project on the humanoid robot 'Optimus' at Tesla's A.I. Day Event in August 2021 [25]. The Optimus is a human-like robot aimed at achieving an increased level of manufacturing automation and is expected to take over dangerous, repetitive, and boring tasks from humans. Tesla aims to integrate this robotic workforce for assembly tasks in automobile manufacturing which is a challenging application of robotic automation [26]. Tesla forecasts that its humanoid robot can become even bigger business than electric automobile manufacturing [27]. Since its inception, the Optimus project has received a lot of attention in the industry and beyond.

With recent developments in humanoids and human-robot interaction technologies [28] [29], it is possible that the vision of resilient and safe human-robot teams may become a reality [30]. It can enable an increased level of automation yet maintaining flexibility. However, several aspects associated with the integration of humanoids into the manufacturing base need to be appraised.

This study explores the usability, commissioning, and associated challenges for manufacturing-related applications. An open-source medical ventilator was used to develop a human-humanoid collaborative (HHC) assembly cell. Usually difficult to be automated tasks through robotic arms are assigned to humanoids.

This article presents:

1. The usability of humanoids for manufacturing automation through a use case and simulations
2. A framework to integrate humanoids in manufacturing applications
3. Challenges and future research directions

## 2. Workforce shortage/skills inadequacy and implications for manufacturing

The worldwide manufacturing ecosystem is facing the challenges of workforce shortage due to: 1) an aging workforce, 2) outdated workforce planning, 3) less efficiency of national education, 4) poor perception of manufacturing among the young generation, and 5) the changing nature of work [31]. It is estimated that the current trends may result in 2.4 million manufacturing jobs unfilled by 2028 [32]. While closing the global skills gap could add US$ 11.5 trillion to global GDP by 2028 [33].

Japan, Italy, Germany, and Sweden present a high manufacturing value added in their national GDPs while exhibiting a percentage of 20.1, 19.7, 18.8, and 17.2 of people above the age of 65, respectively [1]. The *great resignation* is another recent phenomenon showcasing a trend in people to switch jobs more frequently resulting in workforce unavailability [34]. Fierce market competition and expanding wage rates are further adding to the need for automation [35]. Additionally, the pandemic of COVID-19 and social distancing measures have left several question marks for manufacturers on how to be resilient to abrupt measures in such catastrophes [36].

To summarize, globalization, market fluctuations, and skill shortage are the reasons for manufacturers to improve productivity, increase the level of automation, and be more responsive to market dynamics [37][38]. Innovative technologies are increasingly becoming available [39] and promise innovation breakthroughs, including the transformation of industrial ecosystems. However, it is important that the journey of making the best use of emerging technologies is neither straightforward nor short [40].

## 3. Manufacturing automation: where are we and where to go?

Due to the challenge of globalization in the recent decades, many manufacturing companies left the innovation pressure to their suppliers [41]. The geographical distance in research centers and manufacturing facilities resulted into insufficient manufacturing innovation. However, for many years now, there has been a trend to move back the core competencies of companies [42]. For example, the past recent decades have observed innovation in assembly in terms of minimizing wastes and work efficiency only. However, now there is a high interest in assembly automation with robots. Though industrial robots are being used for most of the present-day manufacturing automation [43] but due to their inflexibility, safety and reconfiguration efforts they have not been suitable for assembly [44].

The reliance on human labor in a manufacturing value chain depends upon the design of products, demographic situations, and labor costs. For example, the manufacturing of various products involves operations that are difficult to be automated through robotics. A product that involves a variety of tasks/components and a need of teamwork in its assembly forms a difficult case for mainstream automation [44]. However, with changing demographics, it is becoming important to utilize humans in more value-adding activities, offer them better work conditions, and reduce workload [45]. The recent COVID-19 pandemic also emphasized an unprecedented scalability and reconfigurability of future factories [46].

The collaboration of humans and robots to accomplish a variety of tasks is a long-existing concept [47] but gained traction in recent years [48]. Significant results have been documented and new technologies (to ensure safety and natural human-robot interaction) have been developed [49]. Also, the methods to program and reconfigure robotic systems are evolving [50] [51] (see Figure 1). For example, companies like Boston Dynamics have shown this potential already in real applications [52].

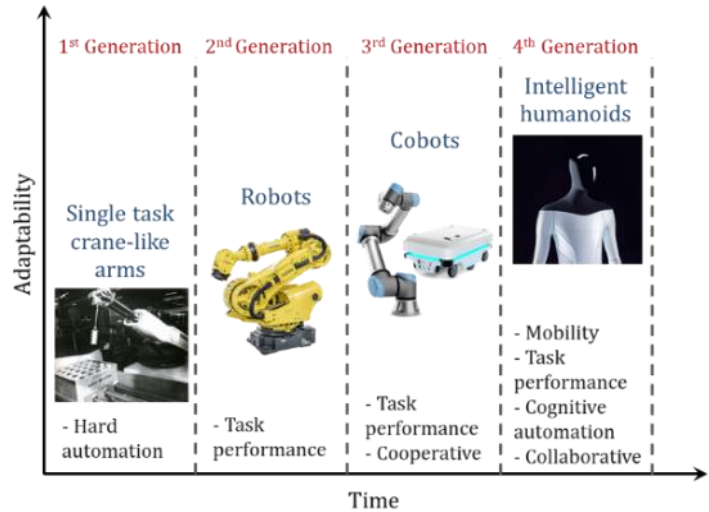

*Figure 1: Evolution of robotic automation in manufacturing.*

Traditionally, manufacturing flexibility and volume are inverse of each other (see Figure 2). But the emerging concept now is that a robot capable of mobility (maneuvering in uncertain environments), performance of complex tasks (e.g. climbing ladders and joining cables), and natural interaction with team members (anticipating and responding) can automate what was not possible before [53]. Such manufacturing automation can redefine the inversely proportional relationship between flexibility and production volume.

If the automation is flexible enough (as humans) and gains a degree of autonomy, it can no longer result in reduced flexibility. Rather it may showcase an interesting ramification where automation is equally flexible as humans but lacks skills of creativity and empathy. As long as these skills are not required in the automation of a certain process then flexibility will no longer be a barrier to automation. It may emerge a new manufacturing paradigm of mass personalization.

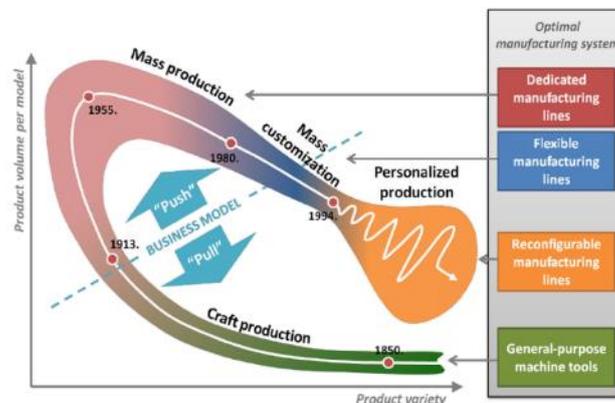

*Figure 2: Different manufacturing systems shaping manufacturing paradigms* [54]*.*

## 4. State of the art in collaborative humanoid robots

A collaborative humanoid robot is desired to engage and collaborate with humans in close proximities and therefore needs to have the capabilities of mobility, task performance, communication, and a degree of autonomy [53]. Several humanoid robots have been documented and presented conceptually or in the labs in literature. A basic definition of a humanoid robot is needed to explore manufacturing applications. For the purpose of this article, a humanoid robot is defined as an autonomous multi-axes reprogrammable electromechanical device that has the capability of mobility, weight carrying, and physical task performance.

4.1. **ASIMO:** Honda Motor Co., Ltd. took up the challenge of developing a mobile humanoid robot named Advanced Step in Innovative Mobility (ASIMO) in 1986 [55] [53]. The first version of ASIMO was introduced in 2000, that displayed two distinct features of humans i.e. bipedal walking and "knowing and learning from humans". Besides a stable walk on uneven and slanted floors, in terms of task performance, the robot was able to perform simple tasks such as handing over a tray, pushing a cart, and was able to pour drink. By using voice and image recognition and plenty of physical characteristics, ASIMO was also able to interact with humans in a natural way [56]. However, ASIMO never made it to actually work on the manufacturing floor, and due to challenges of justifying a profitable business case, the ASIMO project was retired in 2018.

4.2. **Valkrie:** NASA reported a humanoid robot named Valkyrie for human spaceflight endeavors in extraterrestrial planetary settings [57]. Valkyrie stands 1.87 m tall, weighs 129 kg, and approximates a human range of motion. Though NASA had reported other robotics projects in the past, Valkyrie was incorporating much of the learning from those previous projects.

4.3. **HRP-2:** There are barely a few examples where humanoids were translated into serial production. A project named HRP-2 [22] documents the development of humanoids in aircraft assembly (Figure 3). The project involved the integration of a humanoid robot for bracket assembly of the fuselage in A350 aircraft manufacturing at Airbus. The humanoid robots were used in the project to avoid production delays, improve product quality, and perform large quantities of repetitive tasks. Another example is the Glory Factory in Saitama, Japan, which incorporates humanoids to assemble money-handling machines [22]. The past insignificant attention that humanoids could get for manufacturing automation was due to (i) technological immaturity, (ii) justifying a realistic return on investment, (iii) complexity of commissioning, and (iv) difficulties in reconfiguration.

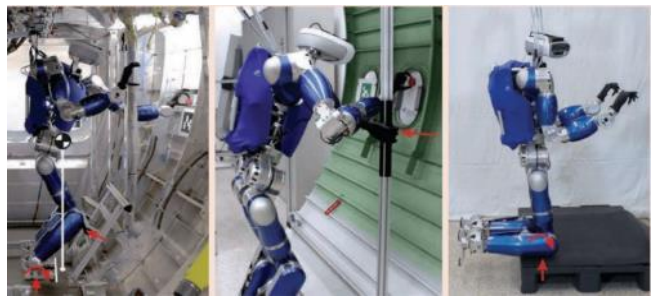

*Figure 3: Humanoid for bracket-assembly of the fuselage in manufacturing of A350 aircraft [22].*

4.4. **Optimus:** The latest development of Optimus (a humanoid robot introduced by Tesla) [25] features a human body-like architecture, a weight of 57 kg, and a height of 1.73m, thus concluding a body mass index (BMI) of 19. With such an unelevated BMI, it has a weightlifting capacity (payload) of 20 kg, but with stretched arms, the payload capacity is reduced to 10 kg. The Optimus incorporates bipedal mobility and has a maximum speed of 8km per hour. The Optimus is expected to have 40 electromechanical actuators of which 12 are in the arms, 2 each in the neck and torso, 12 in the legs, and 12 in the hands (Figure 4).

Optimus is aimed to be equipped with human-like hands staging a high degree of flexibility and dexterity. Additionally, the robot will have a screen on its face to display information usable for cognitive automation. Several proprietary features of Tesla will be incorporated into the robot, such as a full-self-driving computer, autopilot cameras, full suite A.I. tools, neural net planning, auto labeling for objects, simulation capability, etc.

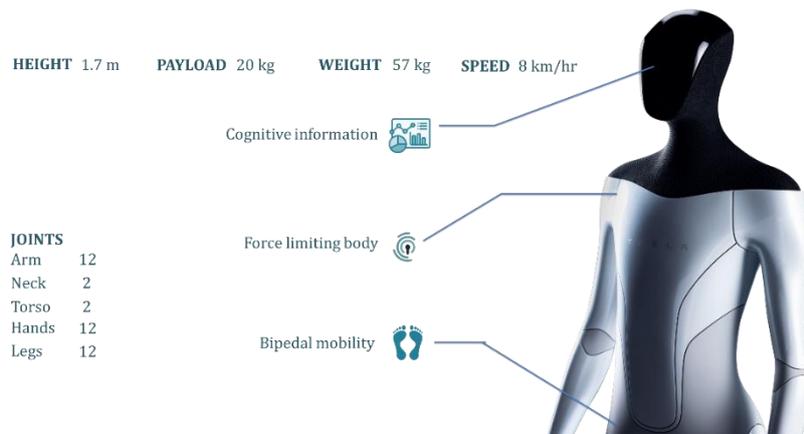

*Figure 4: Technical specifications of humanoid robot introduced by Tesla* [25]*.*

## 5. Research Methodology

For a humanoid robot to engage and collaborate with humans in close proximity needs to have the capabilities of mobility, task performance, communication, and a degree of autonomy [53]. Each of these capabilities is of can be used to These capabilities set forth the evaluation criteria for this study. The paper is a simulation-based investigation of the application of a humanoid in a team setting for the assembly of a medical ventilator.

The manual assembly process of the use case is first investigated for HHC automation potential by decomposing the assembly process into tasks and identifying the automation potential of each task. The tasks, based on their automation potential, are then distributed among humans and robots. It is investigated if HHC carries higher automation potential as compared to conventional robotic arms and its appositeness in teamwork in assembly cells. Additionally, lean manufacturing aspects are investigated.

## 6. Framework to integrate humanoids in factories

Commissioning of humanoids in manufacturing is supposed to offer flexibility, safety, and ease of reconfiguration. Smart manufacturing technologies can be used to reduce the time to design an HHC manufacturing system, validate it and commission it. A structured framework to integrate humanoids (Figure 5) in assembly and other manufacturing applications is presented in the section below:

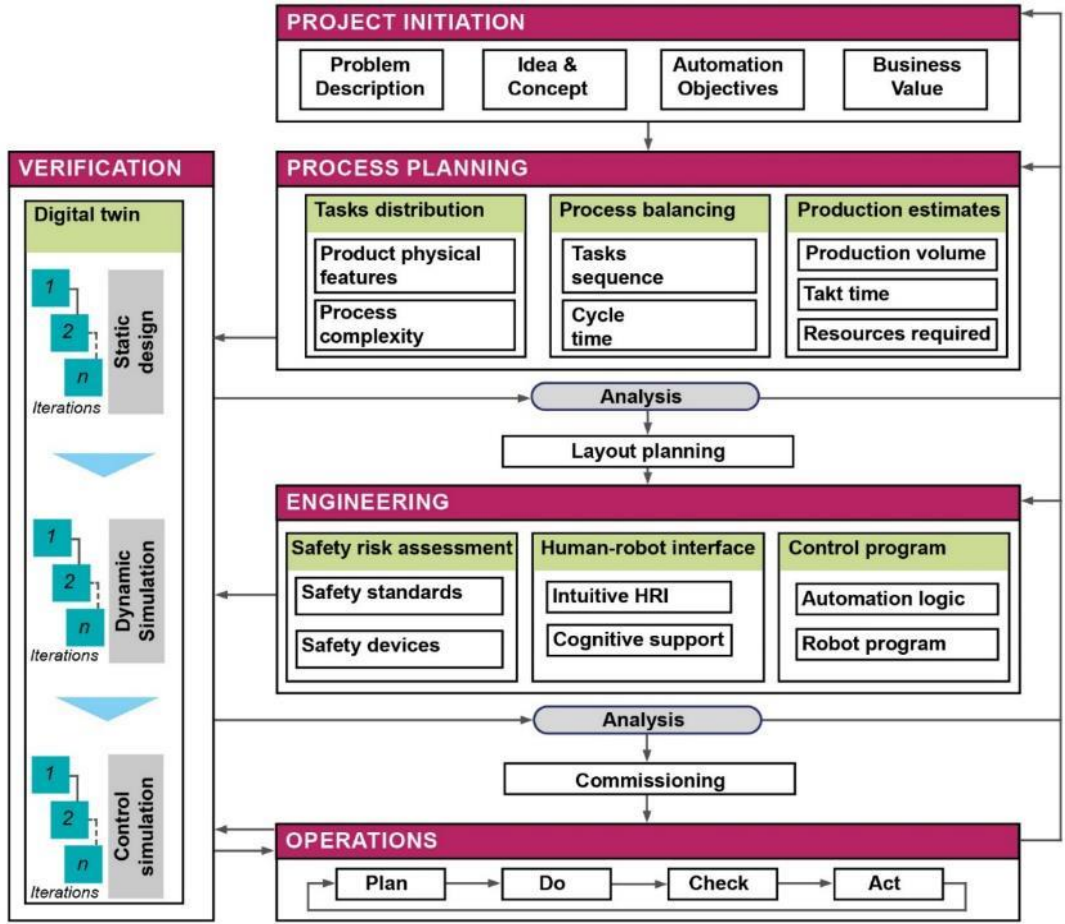

*Figure 10: Framework to integrate humanoids in manufacturing.*

The framework starts by defining the problem that HHC automation can potentially address. Automation objectives are documented, and an initial concept of the automation system is developed. It helps to make an initial financial assessment. The deliverable from this step is a list of objectives and concepts of humanoid automation. Deliverable also acts as a success criterion for the automation project.

The suitability of a task for HHC is based upon the physical characteristics of the parts/components (shape and material characteristics), parts presentation (structured presentation to the robot), variability (of tasks), and risk of injury to humans around [19]. Various structured methods to assess product complexity for robotic automation have been presented in various literature, such as [72][73][74][75].

## 7. Case study and simulation

This section presents the case of automating a manual manufacturing assembly cell using humanoids. The product used to study the usability of HHC automation is an open-source medical ventilator PB560 [58] by Medtronic (Figure 5). With a workforce of more than 90,000 people across 150 countries, Medtronic stands out as a leading company in medical device manufacturing. At the peak of the COVID-19 pandemic, to address the shortage of medical ventilators, the company open-sourced one of its ventilators (i.e. PB560) [59]. The move included open sourcing of the technical drawings, associated manufacturing processes, electrical diagrams, and the required quality tests.

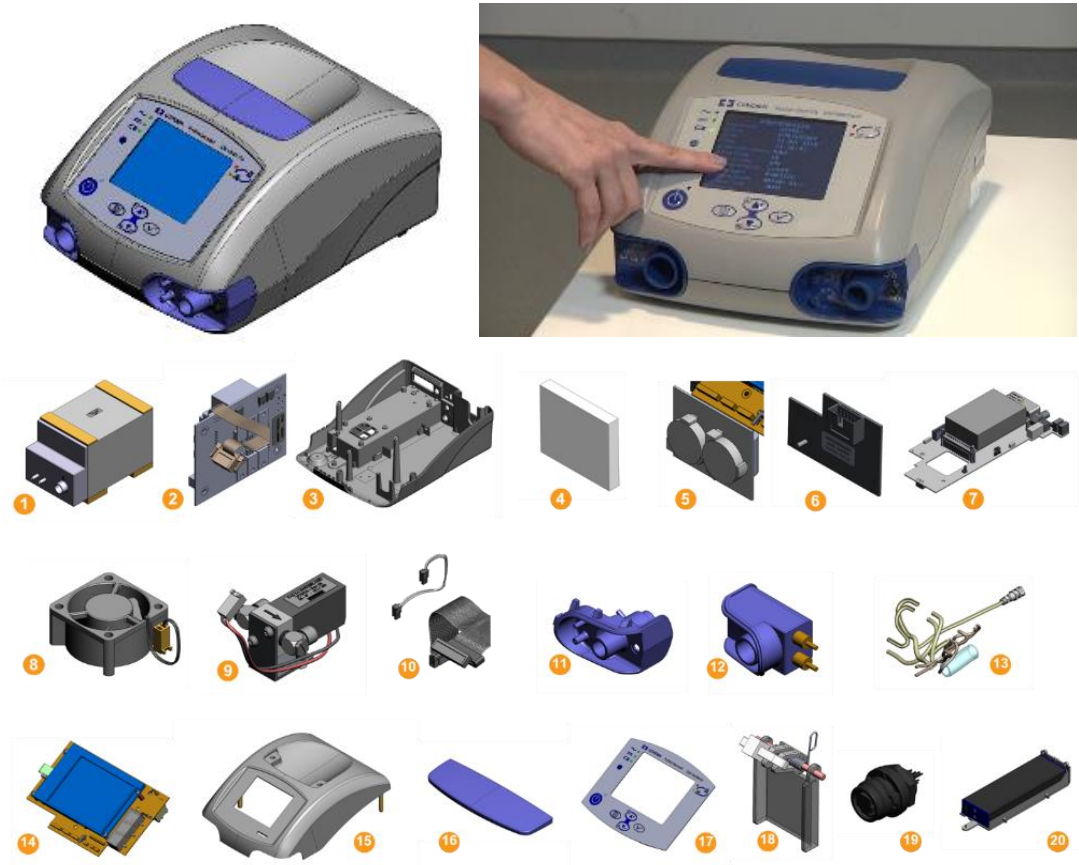

*Figure 5: Open source PB560 mechanical ventilator by Medtronic.*

The PB560 ventilator consists of 20 unique parts and sub-assemblies (Figure 6). The assembling of these parts and subassemblies into a finished product consists of twenty tasks. The precedence sequence of these tasks is shown in Figure 10. In addition to assembly tasks, there are additional tasks of quality testing and material handling involved. The product establishes a strong use case to assess HHC usability as it involves the assembly of medium to complex tasks and is open source (can be reproduced).

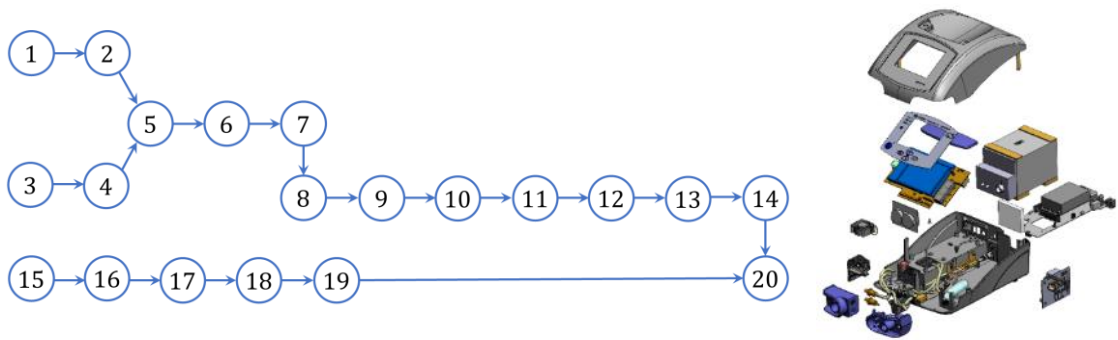

*Figure 6: Assembly precedence sequence of PB560 medical ventilator.*

### 7.1. HHC automation planning

This section assesses HHC automation to assemble the case product. The evaluation of tasks for humanoid automation contrasts with conventional robotic arm-based automation as humanoids have: i) two arms; ii) mobility; iii) collaboration characteristics with humans; and iv) safety features. The robot used in this study is Tesla Optimus, which has six degrees of freedom, a payload capacity of 10 kg, and a reach of 1,000 mm for each arm. The default five fingers human-hand gripper is used with the robot.

*Table 1: Evaluation of assembly tasks for assignment to robot [36].*

| Name | Assembly attributes | Rating 1/0 (1= robot can do; 0= robot can't do) | Resource |
|---|---|---|---|
| Task | Physical shape (P) | If a robot can handle the part | If (S && F && M && J && Sf == 1, "Robot"; else "Human") |
|  | Part feeding (F) | If parts have a structured presentation |  |
|  | Mounting (M) | If no precise adjustment is needed |  |
|  | Joining (J) | If joining a task is easy for the robot |  |
|  | Safety (S) | No safety risk if performed by the robot |  |

A simplified method for task distribution is shown in Table 1. The assembly process of PB560 is decomposed into tasks, and each task is evaluated for its automation potential (Table 1). A task, if automated, doesn't carry any challenge in any of the evaluation criteria (i.e., handling, feeding, mounting, joining, and safety), then presents a high potential for automation. Though many of the tasks can be automated, a process balancing is needed to optimally automate the tasks.

*Table 2: Evaluation of tasks of the use case assembly for automation potential.*

|  | T1 | T2 | T3 | T4 | T5 | T6 | T7 | T8 | T9 | T10 | T11 | T12 | T13 | T14 | T15 | T16 | T17 | T18 | T19 | T20 |
|---|---|---|---|---|---|---|---|---|---|---|---|---|---|---|---|---|---|---|---|---|
| Part (P) | 0 | 0 | 0 | 0 | 0 | 0 | 0 | 0 | 0 | 1 | 1 | 1 | 1 | 0 | 0 | 0 | 0 | 0 | 0 | 0 |
| Feeding (F) | 0 | 0 | 0 | 0 | 0 | 0 | 0 | 0 | 0 | 1 | 1 | 1 | 1 | 0 | 0 | 0 | 0 | 0 | 0 | 0 |
| Joining (J) | 0 | 0 | 0 | 0 | 0 | 0 | 0 | 0 | 0 | 0 | 0 | 0 | 0 | 0 | 0 | 0 | 0 | 0 | 0 | 0 |
| Mounting (M) | 0 | 1 | 0 | 1 | 0 | 0 | 0 | 0 | 0 | 1 | 1 | 1 | 1 | 0 | 0 | 1 | 0 | 1 | 0 | 0 |
| Safety (S) | 0 | 0 | 0 | 0 | 0 | 0 | 0 | 0 | 0 | 0 | 0 | 0 | 1 | 0 | 0 | 0 | 0 | 0 | 0 | 0 |
|  | R | H | R | H | R | R | R | R | R | H | H | H | H | R | R | H | H | H | R | R |

### 7.2. HHC assembly system design

The proposed HHC assembly cell comprises five workstations, of which two are operated by humanoids, and three are manual. Once an initial task identification has been performed for robotic or manual tasks, a refined process description is needed to balance the process (Table 3). Lean assembly balancing involves the assignment of tasks to workstations and ensuring that no workstation becomes a bottleneck, minimizing idle time and compliance with the takt time. A schematic of the proposed assembly cell is shown in Figure 7.

Calculations to set up a lean HHC assembly cell are given below:

    i.   *Total processing time for one unit = $T_P$ = 803 sec*

ii. Time in a day = $T_D$ = 27,000 sec
iii. Required units in a day = N = 167 units
iv. Takt time = $T_K$ = $\frac{T_D}{N}$ = $\frac{27,000}{224}$ = 161 sec
v. Required number of workstations N .W. = 5
vi. Total automated tasks = $T_H$ = 321 sec = 40%
vii. Total manual tasks $T_M$ = 482 sec = 60%

*Table 3: Tasks distribution in HHC assembly cell and process balancing.*

| Workstations | Resource | Tasks | Task time | Time left |
|---|---|---|---|---|
| Workstation 1 | Robot | 1, 2, 3, 4, 5, 6 | 160 | 0 |
| Workstation 2 | Human | 7, 8, 9, 10 | 161 | 0 |
| Workstation 3 | Human | 11, 12, 13 | 161 | 1 |
| Workstation 4 | Robot | 14, 15, 16, 17 | 160 | 0 |
| Workstation 5 | Human | 18, 19, 20, 21 | 161 | 1 |

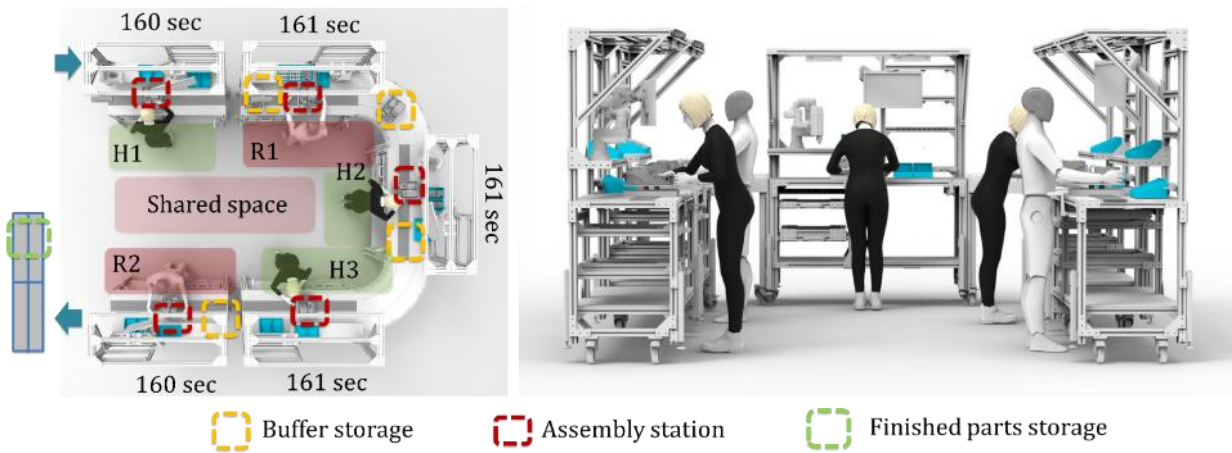

*Figure 7: Assembly precedence sequence of PB560 medical ventilator.*

### 7.3. Simulation development

Since the article explores the usability of humanoids in assembly cells, high-fidelity simulations appeared to be a useful method for making the assessment. The digital twin method starts by creating a static model of the assembly cell, then developing a continuous event-based simulation for kinematic visualization and cycle time estimates, and exporting the results to a stochastic simulation for production and material flow estimates.

The simulation helped in the following:

1. Validation of robot kinematics to perform assigned assembly tasks
2. Design of the assembly cell and process balancing

3. Cycle time estimates for robotic tasks
4. Economic estimates of using humanoids in assembly cells

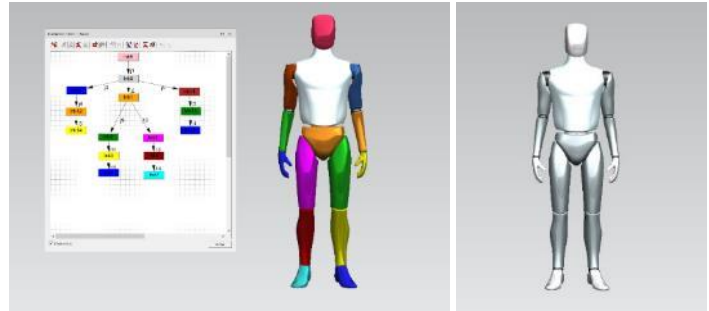

*Figure 8: Creating a kinematic robot model in simulation.*

A static CAD model of a humanoid robot was used from open-source GrabCAD. The model was then imported into Tecnomatix Process Simulate (TPS) in J.T. (Jupiter Tessellation) format. TPS is a continuous simulation software and can be used for robotic process modeling. The kinematics of the humanoid were created in TPS (Figure 8). Five assembly workstations were used in the cell, of which three were designed as manual, and two were designed as robotic workstations. The robotic and manual assembly tasks were modeled in the simulation.

The layout of the workstation was also designed considering the arm reach of humans and robots. The design of the workstation is usable both for humans and robots so that they can exchange the workstations if desired. The simulation also helped to identify possible collisions in the given robot trajectories. The robot paths were also optimized to avoid any unnecessary movements, reduce cycle time, as well as any possible collisions.

The results from the continuous simulation are exported to an event-based simulation for economic and throughput estimates. Different scenarios are created to induce product variants, changeover time variation, mean time to failures, and setup time (including the charging time of the robots). The results are shown in the figure below.

## 8. Results and findings

The product used in the case study is a complex product, but still, seventy percent of the tasks appear to be automatable. These seventy percent of tasks correspond to seventy-five percent of the total assembly time. Therefore, the use of robots can reduce up to seventy-five percent of manual assembly time. However, due to task precedence and process balancing, sixty percent of the total tasks have actually been automated in the simulation study. Additionally, material handling tasks have also been automated, which account for 60 minutes in total in a shift.

Different process variabilities are introduced in the simulation model, such as cycle time, collisions, incorrect parts, and delays. Additionally, different products are run with different cycle times and task distributions. A total of 50 variants are simulated with different cycle times resulting in varying task distribution strategies and automation potential. It is observed that the HHC assembly cell can effectively automate most of the products.

Additional results are documented below:

## 8.1. The implication of safety standards for humanoids

Presently, all manufacturing-related applications of stationary collaborative robots need to comply with ISO-15066 safety specifications [60]. The ISO-15066 requires that the robot and the robot system (including any additional hardware, fixtures, grippers, and tools) must be equipped with a safety-rated monitored stop, power, and force limiting body, speed and separation monitoring, and hand guiding feature (see Figure 9). For any application of the robot, a safety risk assessment needs to be carried out before putting the robot in action.

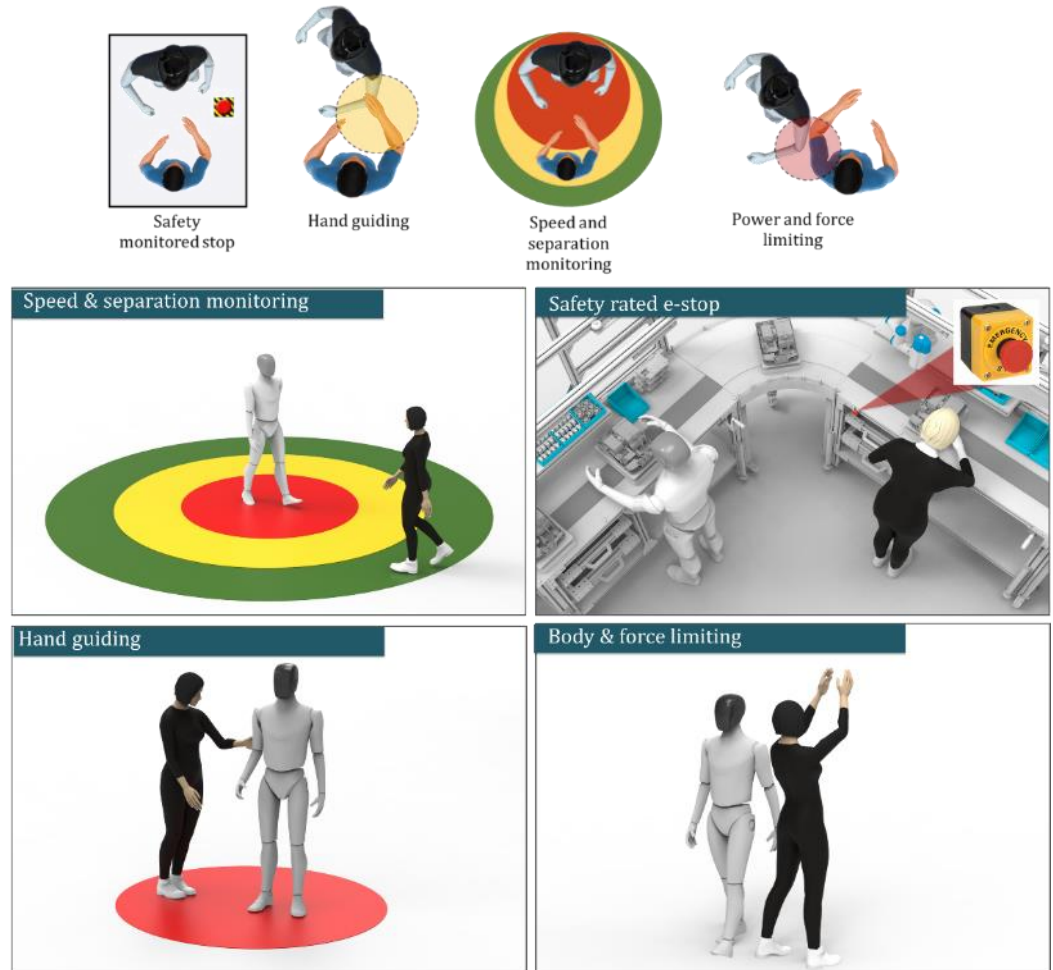

*Figure 9: Illustrations of different forms of human-humanoid interaction on the manufacturing floor in the light of ISO-15066.*

Complying with these requirements would not be a challenge for humanoids. However, in close proximity to humans, the allowable speed of the robot is under 250mm/sec. This permissible speed is lower than an average human arm speed which can challenge compliance with takt time. In addition to ISO-15066, humanoids need to comply with safety standards for mobile robots, namely ISO-19649. Asp per ISO-19649, humanoids may be categorized as IMR type-C (mobile platform with manipulator attachment). A safety description of each form of human-robot cooperation as per ISO-15066 is given below:

I. The complete body of the humanoid must be equipped with the **power and force limiting** feature.
II. A **safety-rated stop** must be available at arm's length for the operator.

III. **Speed and separation monitoring** must be enabled for to robot to detect the presence and distance of humans from it and change its speed depending upon the distance from a human.
IV. The feature of **hand guiding** must be available to let a human manually define the robot's trajectory online.

Machine learning (ML) methods can help to optimize robot trajectories and minimize the risk of collisions. ML can be used to make the robot learn from historical data and optimize its performance in emerging situations. Additionally, floor markings can be used, designating areas with absolute human, absolute robot, and hybrid spaces.

## 8.2. Forms of humanoid-human collaboration:

HRI is an evolved form of human-machine interaction. Recent scientific publications outline five forms of human-robot collaboration [61] [62]. The categories are based on time and space sharing between humans and robots in a team scenario. In the case of HHC, a description of these forms is given below (see Figures 10):

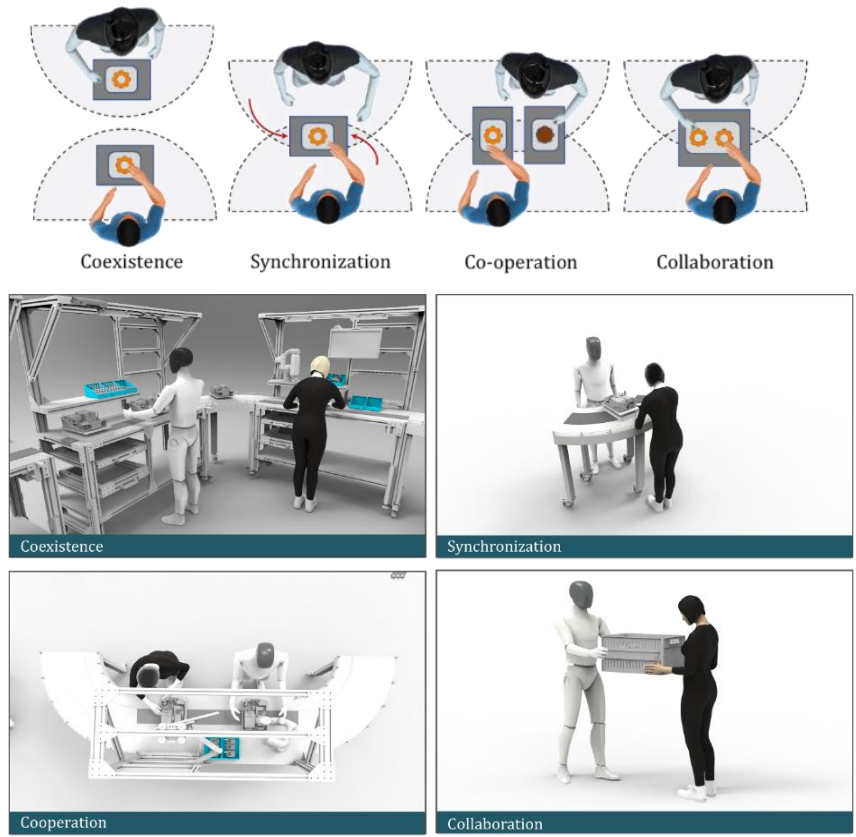

*Figure 10: Forms of human-humanoid collaboration.*

**Coexistence:** Humans and robots can share the workspace; however, they have individual independent tasks. Due to limited to no direct interaction between humans and robots, this form of interaction can be useful for lean assembly cells.

**Synchronized:** In this form of collaboration, humans and robots can work on the same workpiece but in turns. Therefore only one of them is active at a given time. It can be useful in applications where some tasks require higher quality achievable through automation or if the task may cause safety risks. For example, at an assembly table, where a robot can perform the lifting of heavy objects, and humans can perform dexterous tasks.

**Cooperation:** Humans and humanoids can have a common workspace, and they both are active at the same time. However, they have separate tasks.

**Collaboration:** It is the highest level of collaboration, possessing the most risk and most difficulty to achieve. It highlights an active collaboration to accomplish a difficult task. For example, they both carry a load, or the operator holds a part, and the bot performs a task on it.

### 8.3. Digital twin to design, develop and operate HHC work cells

A hybrid simulation is developed where equation-based models are generated that represent the resources (humans, robots, material, etc.). A model for simulation-based design is presented in Figure 11. First, the interaction between models (resources and stocks) is studied through dynamic time-lapsed continuous simulation. The humanoid and human tasks were executed in this simulation for detailed process visualization and cycle time estimation. The process was also balanced to avoid any waiting time.

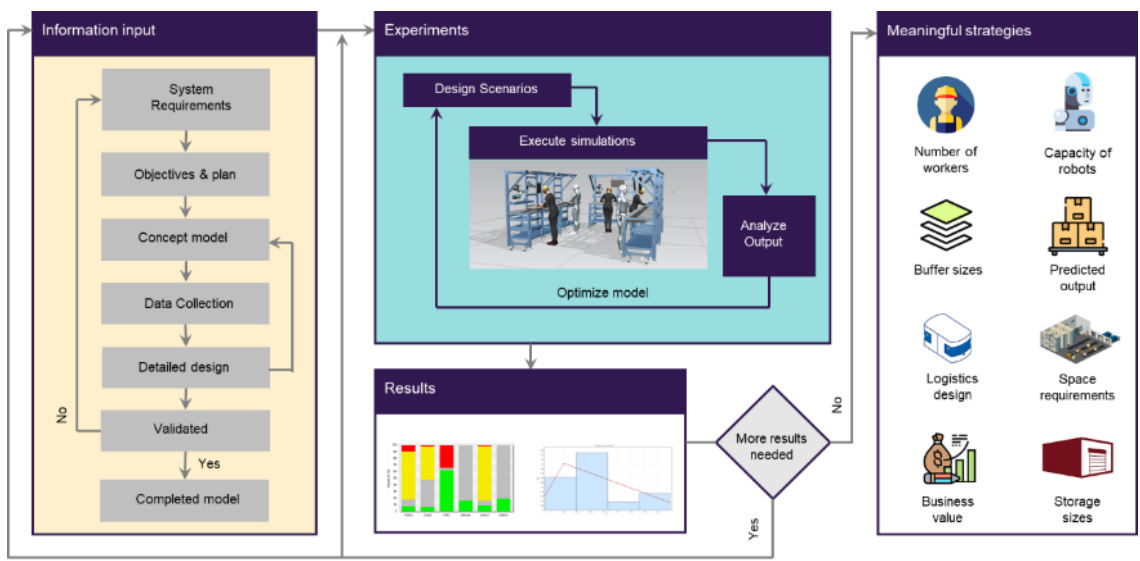

*Figure 11: Model to utilize simulations and digital twins in the development of HHC cells.*

The results were then exported to discrete event simulation (DES) to study the variability in time taken to carry out the activities. DES is useful for studying the impact of resource availability and production volumes and estimating the lead time to fulfill customer demand [63].

The rationale behind using two types of simulations has also been proposed by [64], as C.S. accounts for the time-dependent dynamic behavior of the system while DES includes stochastic effects. By using the two simulations, the advantages are combined while minimizing the shortcomings of each.

The same simulation model can then be extended to combine the automation program of the assembly cell, robot programming, and cloud data communication. It can make it easy to commission the cell and reconfigure it.

### 8.4. Possible application areas of humanoid robots

Robots are considered suitable for repetitive tasks requiring a low degree of ergonomics. Therefore, HHC can be useful for the following:

**Material handling:** The lean literature takes material handling as a non-value-adding activity that must be minimized but is often unavoidable. Manufacturers always endeavor to reduce material handling, but when it becomes inevitable, different strategies are adopted, such as conveyor lines, AGVs, and mobile robots. Due to the need for flexibility, mobile robots have gained attraction for this purpose in recent years.

**Pick and place tasks:** Pick, and place is another possible application of humanoids. It may involve picking the parts of interest from a location, moving them to the place of action, and then placing the parts in the required orientation. Pick and place is considered an ideal task for automation, but the physical characteristics of the material being handled can make it difficult for robots.

**Assembly and screws driving:** Many industrial products have extensive screw driving and other fastening applications. For example, a typical wind turbine has >6,000 bolts. A phone is held together with about 75 fasteners, a car with 3,500, and a jet plane with 1,500,000. At the same time, screw driving is a repetitive process with minimum to no variations (easy to automate) from one screw to the next.

## 9. Research recommendations

This section presents the topics that can be considered as research directions for the enablement of humanoid human teams.

### 9.1. Adaptability and robot teaching

Flexibility is an increasingly needed property of a manufacturing system. Flexibility may be required due to product design changes, updated company strategy, or market fluctuations. In any such case, the manufacturing system is required to be modified or reconfigured to align with a new strategy. In any such case, the robots may be required to be reprogrammed. The programming interfaces must be simple and intuitive. It must not take more than what a human team in assembly takes. Arm control features and techniques to learn by watching a task can speed up the process. Another technology is the digital twin. A digital twin can facilitate reinforcement learning so that the robot can learn the tasks from simulations.

### 9.2. Safety implications

Ensuring the safety of human workers will be a primary challenge for system designers and developers. Though this is the use case example, there is no direct human-robot interaction. Complying with ISO-5066 will remain a challenge because the standard requires that a safety risk assessment needs to be performed every time there is a hardware change in the robot system, which includes the layout, the robot's trajectories, and any devices attached to the robot. But this will bring challenges to maintaining manufacturing flexibility. During material transportation, the robot and humans may get even closer posing more risks.

### 9.3. Human-robot interaction

In a lean assembly cell, the team members are supposed to work in teams, share the workload, and maintain the work pace. To get closer to this level of collaboration, fluid interaction between humans and robots is needed. Some of the modern techniques to achieve fluid interaction can be voice recognition, gesture recognition, wearable devices, or brain wave control. For needed reliability in manufacturing applications and for the safety of human workers, wearable devices and hard buttons can be an optimal way to achieve interaction.

## Conclusion

Achieving a high degree of flexibility in manufacturing automation is still an unfulfilled desire of manufacturers. Intelligent humanoid robots can push the boundaries of manufacturing automation from rigid conventional automation to adaptive, resilient automation. Humanoids can increase the level of automation while maintaining flexibility. The developments in humanoids can revolutionize the concepts of robotic automation especially in assembly. Reconfigurability and teaching the robot for changeovers can be significant challenges that are difficult to deal with the present-day robotic technologies. Digital twin is an emerging technology that can facilitate to address the complexity of humanoid-human automation. The humanoid robots must come up with innovative approaches for reconfiguration.